\documentclass{article}

\PassOptionsToPackage{numbers, sort&compress}{natbib}

\usepackage{arxiv}
\usepackage{amsmath}
\usepackage{amsthm}
\usepackage[ruled]{algorithm2e}
\usepackage{graphicx}

\usepackage[utf8]{inputenc} %
\usepackage[T1]{fontenc}    %
\usepackage{hyperref}       %
\usepackage{url}            %
\usepackage{booktabs}       %
\usepackage{amsfonts}       %
\usepackage{nicefrac}       %
\usepackage{microtype}      %
\usepackage{xcolor}         %
\usepackage{bbding}

\definecolor{citecolor}{HTML}{0071BC}
\definecolor{linkcolor}{HTML}{ED1C24}
\hypersetup{colorlinks=true, linkcolor=linkcolor, citecolor=citecolor,urlcolor=black}

\usepackage{bm}
\usepackage{placeins}
\usepackage{afterpage}

\title{FILM: How can Few-Shot Image Classification Benefit from Pre-Trained Language Models?}

\author{%
  \textbf{Zihao Jiang$^{1,}$\thanks{Equal contribution. This work was done when Yunkai was visiting Qing Yuan Research Institute.}
  \quad
  Yunkai Dang$^{2,*}$
  \quad
  Dong Pang$^3$
  \quad
  Huishuai Zhang$^4$
  \quad
  Weiran Huang$^{1,}$\thanks{Correspondence to Weiran Huang (weiran.huang@outlook.com).}}\\[0.3cm]
  $^1$ Qing Yuan Research Institute, Shanghai Jiao Tong University\\
  $^2$ College of Intelligence and Computing, Tianjin University\\
  $^3$ School of Mechanical Engineering, Shanghai Jiao Tong University\\
  $^4$ Microsoft Research Asia
}

\begin{document}

\maketitle

\begin{abstract}

Few-shot learning aims to train models that can be generalized to novel classes with only a few samples. 
Recently, a line of works are proposed to enhance few-shot learning with accessible semantic information from class names.
However, these works focus on improving existing modules such as visual prototypes and feature extractors of the standard few-shot learning framework.
This limits the full potential use of semantic information.
In this paper, we propose a novel few-shot learning framework that uses pre-trained language models based on contrastive learning.
To address the challenge of alignment between visual features and textual embeddings obtained from text-based pre-trained language model, we carefully design the textual branch of our framework and introduce a metric module to generalize the cosine similarity.
For better transferability, we let the metric module adapt to different few-shot tasks and adopt MAML to train the model via bi-level optimization.
Moreover, we conduct extensive experiments on multiple benchmarks 
to demonstrate the effectiveness of our method.

\end{abstract}
\section{Introduction}
Deep neural networks \cite{AlexNet,VGGNet,GoogLeNet,ResNet} have achieved remarkable success in many fields. 
However, training deep neural networks requires a large number of labeled data, which can be expensive and time-consuming to obtain.
For instance, in medical imaging, obtaining labeled data requires expert radiologists to annotate images. 
This limits the application of deep learning models in real-world scenarios. 
In contrast, humans possess the ability to recognize and classify objects of unseen categories with only a few examples. 
This highlights the potential value of few-shot learning \cite{FSL1, FSL2, FSL3, FSL4}, where models are trained on base classes and can be generalized well to novel classes with limited amounts of samples. 

Previous works mainly focus on image classification tasks, and most of them adopt the meta-learning paradigm \cite{MatchingNet,ProtoNet,MAML,RelationNet,DeepEMD}.
Recent works consider leveraging additional information from other modalities such as text to enhance the performance of few-shot learning.
In particular, some methods \cite{AM3,KTN,TRAML} adopt static word embedding models (e.g., GloVe \cite{GloVe}) to extract textual representations of class names and use them to adjust visual prototypes or classifiers. 
With the appearance of general language models such as BERT \cite{BERT} and GPT \cite{GPT}, another line of works \cite{VS-Alignment,SP} adopt public pre-trained language models (PLMs) to extract more comprehensive semantic information from class names.
However, these works still focus on improving existing modules of the standard few-shot learning framework (e.g., visual prototypes and feature extractors), which confines the full utilization of powerful PLMs in few-shot learning.

\begin{figure}[t]
    \centering
    \includegraphics[width=0.8\textwidth]{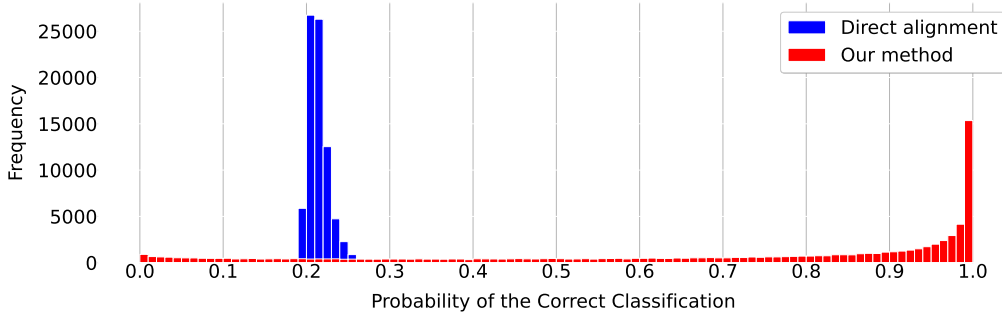}
    \caption{Frequency histogram of probability that each sample image is classified to true label. 80000 samples on novel classes of \textit{mini}ImageNet dataset are collected with 5-way 5-shot setting. For direct alignment, we directly align visual features and textual embeddings extracted by text-based pre-trained language model from class names with cosine similarity. The horizontal axis reflects the probability that each sample image is classified to its true label, which is output by the model. The vertical axis represents the total number of samples in each probability interval.}
    \label{fig:motativation}
\end{figure}

Inspired by the success of vision-language models \cite{CLIP,ALIGN} trained by contrastive learning, we explore the idea of aligning visual features and textual embeddings for few-shot image classification in this paper, where textual embeddings are extracted by a public PLM from class names following the setting of \cite{VS-Alignment,SP}.
However, there are two main factors making this alignment challenging.
Firstly, 
    unlike vision-language models that have sufficient pairs of image and textual descriptions available for model training, we only have the class name of each image instead of a rich description.
Secondly, 
in contrast to vision-language models where both visual and textual encoders are learnable to align embeddings, our textual encoder inherits from a puublic PLM trained on uni-modal text data.
This leads to totally different structures of textual embedding spaces and thus makes the alignment between visual and textual features difficult.
For instance, if we directly align visual features and textual embeddings, the probability\footnote{Here probabilities mean the elements outputted by softmax function.} of a sample image being assigned to its true label is extremely low (see blue bars in Figure~\ref{fig:motativation}). 
This indicates that the visual feature of an image is hard to approach the corresponding text embedding of its true label.

In this paper, we propose a novel framework (Figure~\ref{fig:overall}) to boost few-shot learning by means of public PLMs.
To bridge the gap between visual and textual modalities, we carefully design a textual branch of our framework and introduce a metric module to measure the similarity between visual and textual embeddings. 
The textual branch first incorporates class labels into our hand-crafted prompt template containing a $\rm [MASK]$ token and then inputs the filled sentence to a PLM. 
The PLM transforms the input sentence into a hidden vector sequence
and the final textual embedding is extracted from the vector corresponding to the $\rm [MASK]$ token. 
Meanwhile, the visual feature is obtained by a standard visual encoder.
After that, we compute the similarities between visual features and textual embeddings through the proposed metric module, and send them into the contrastive loss.
For better transferability on novel classes, we let the metric module adapt to different few-shot tasks and adopt Model-Agnostic Meta-Learning (MAML) \cite{MAML} to train the model via bi-level optimization.
Moreover, we conduct extensive experiments on multiple benchmarks to demonstrate that the proposed method significantly outperforms the state-of-the-art few-shot learning methods based on PLMs.

The main contributions of this paper can be summarized as follows.

\begin{itemize}

    \item We propose a novel few-shot learning framework that leverages semantic information extracted by a pre-trained language model based on contrastive learning.
    \item We carefully design a textual branch of the framework and introduce a metric module to generalize the similarity measure.
    \item The metric module is designed to be adaptive to different few-shot tasks for better transferability, and MAML is adopted to train the model via bi-level optimization.
    \item We conduct extensive experiments on multiple benchmarks with different domains to demonstrate the effectiveness of our method.
\end{itemize}

\section{Related Work}

\textbf{Few-shot Learning.}
In general, few-shot learning methods are mainly divided into two categories: metric-based methods and optimization-based methods.
Metric-based methods aim to map samples into an appropriate embedding space on the basis of certain distance metrics. Most previous methods use task-agnostic distance metrics, e.g., cosine similarity distance \cite{MatchingNet}, Euclidean distance \cite{ProtoNet}, CNN relation module \cite{RelationNet}, and Earth Mover’s Distance \cite{DeepEMD}.
Additionally, several methods \cite{Tapnet,CTM,TEAM,ye2020few,DSN} involve learning task-specific distance metrics, which can be adjusted for different tasks.
Optimization-based methods \cite{MAML,LEO,MTL,MetaOpt} aims at learning optimal initial model parameters on base classes and quickly fine-tune them on novel classes with a few support examples.
Our paper generalizes the similarity measure by the proposed metric module, and uses MAML \cite{MAML} to train the model.

\textbf{Few-shot Learning with Semantic Information.}
Recent works on few-shot learning start to utilize semantic information from class labels to enhance few-shot learning.
AM3 \cite{AM3} proposes an adaptive modality mixture mechanism to model prototype representation as a combination of visual features and language semantic features.
KTN \cite{KTN} learns classifiers by fusing visual information and knowledge information acquired from a knowledge graph and word embeddings with a semantic-visual mapping network based on Graph Convolutional Network \cite{GCN}.
VS-Alignment \cite{VS-Alignment} introduces a contrastive alignment between visual and semantic features as an additional objective.
Semantic Prompt \cite{SP} considers semantic information as prompts to tune the ViT \cite{ViT} feature extractor.
All these methods leverage semantic features as auxiliary information to adjust visual prototypes, classifiers, or feature extractors. 
In contrast, we propose a new few-shot learning framework to directly align visual and textual embeddings via contrastive learning.

\textbf{Contrastive Learning.}
Contrastive learning is a popular method in self-supervised representation learning.
It learns representations by pulling positive samples close and driving negative samples away from them in the latent embedding space with a contrastive loss. 
A set of previous works have shown the excellent performance of contrastive learning in computer vision \cite{MoCo,MoCov2,SimCLR} and natural language processing \cite{WordAlignment,LanguageModeling,MachineTranslation} tasks. 
Furthermore, recent works \cite{ConVIRT,CLIP,ALIGN,Flamingo,VS-Alignment} apply contrastive learning to multi-modal settings by aligning image-text pairs in the embedding space. 
Our work introduces contrastive learning to few-shot learning, and proposes a learnable metric module to make aligning visual features and textual embeddings possible.

\section{Problem Definition}

Few-shot learning involves two disjoint class sets: a base class set $\mathcal{C}_{base}$ classes and a novel class set $\mathcal{C}_{novel}$ classes. 
Sufficient labeled samples are provided for each base class, while abundant unlabeled samples and only a few labeled samples are provided for each novel class.
Few-shot learning targets at classifying unlabeled samples from novel classes through training on all the given labeled samples.
Previous works usually formulate the few-shot learning problem as $N$-way $K$-shot classification, which denotes a classification task among $N$ classes with $K$ labeled samples available for each class.
In addition, given a fixed pre-trained language model, we use bimodal contrastive learning to leverage the semantic information extracted by it.
Concretely, for each embedded sample image $z$ and $N$ embedded class labels $\{t_1,t_2,\dots,t_N\}$ in a $N$-way $K$-shot classification task, contrastive learning adjusts the embedding space through the following widely-used contrastive loss \cite{oord2018representation,SimCLR,MoCo,MoCov2} (using cosine similarity as an example): 
\begin{equation}
\label{contrastive_loss}
\mathcal{L} = 
-\log\frac
{\exp(z\cdot t_{+}/\tau)}
{\sum^N_{i=1}
\exp(z\cdot t_i/\tau)},
\end{equation}
where $t_{+}$ is the embedded true label of the sample image and $\tau$ is a temperature hyper-parameter.

Meta-learning paradigm \cite{MatchingNet} is commonly used to solve the few-shot learning problem, 
which trains and evaluates the model with the episodic mechanism.
The standard meta-learning paradigm contains two stages: meta-training and meta-testing.
In each episode of the meta-training stage, a $N$-way $K$-shot $M$-query classification task $\mathcal{T}=(\mathcal{S},\mathcal{Q})$ is constructed with samples from the base classes. 
We first randomly select $N$ classes from $\mathcal{C}_{base}$ as $\mathcal{C}_{\mathcal{T}}$.
For each class, we randomly sample $K$ support images and $M$ query images.
Then we form the support set $\mathcal{S}=\{(x_i,y_i)|y_i\in\mathcal{C}_{\mathcal{T}},i=1,2,\dots,N\times K\}$ and the query set $\mathcal{Q}=\{(x_i,y_i)|y_i\in\mathcal{C}_{\mathcal{T}},i=1,2,\dots,N\times M\}$ with the support images and the query images respectively, where $x_i$ is the $i$-th sample image and $y_i$ is the class label of $x_i$. 
To learn an appropriate embedding space, bi-level optimization is performed on $\mathcal{S}$ and $\mathcal{Q}$ respectively, utilizing a contrastive loss.
In each episode of the meta-testing stage, a classification task is built on the novel classes in a similar way.
The support set is formed with a few label samples, while the query set is sampled from the unlabeled samples.
After adapting to the novel classes by minimizing the contrastive loss on the support set, the model is used to predict class labels for the sample images in the query set.

\section{Method}
\begin{figure}[t]
    \centering
    \includegraphics[width=0.9\textwidth]{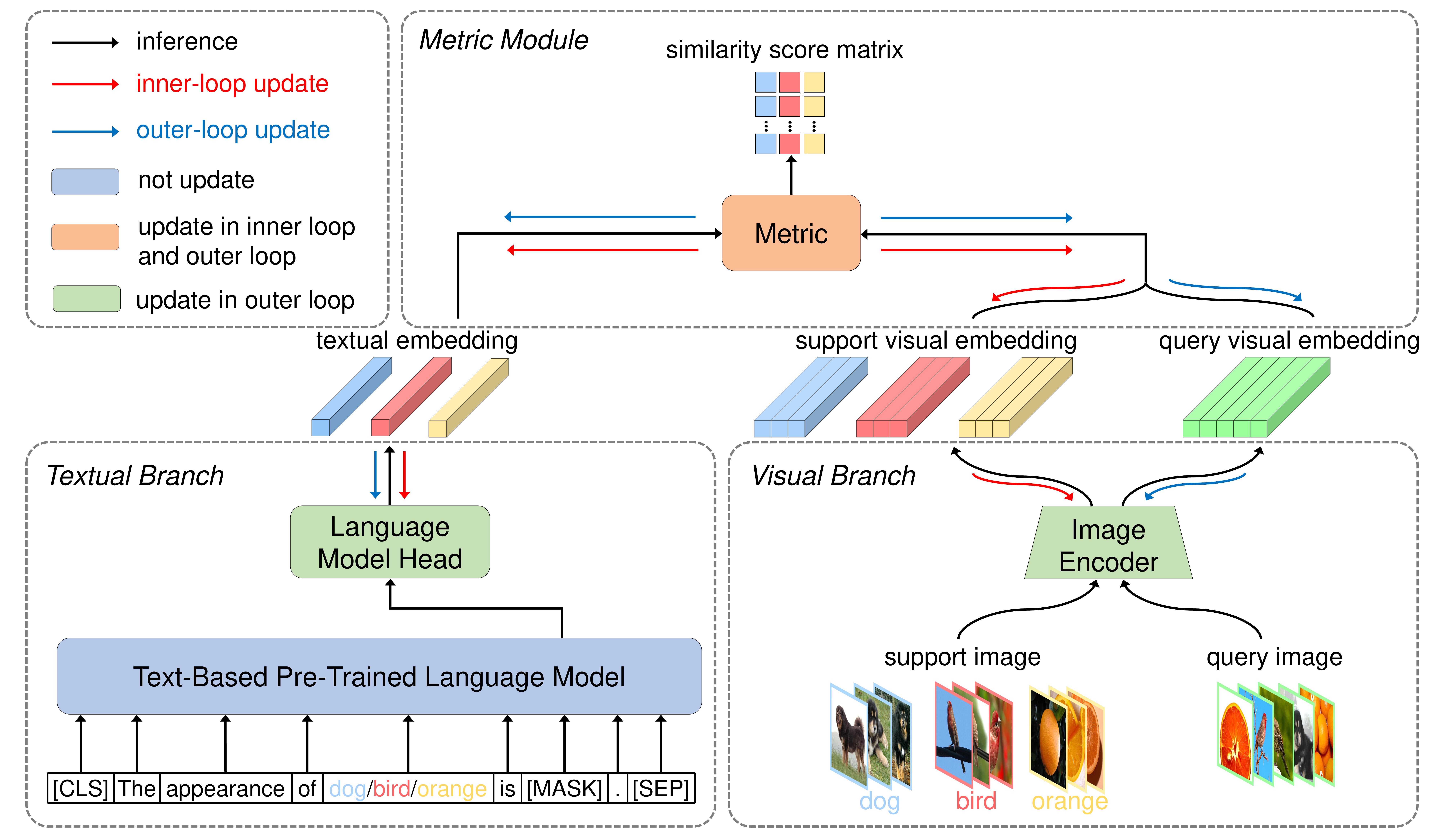}
    \caption{
    The overview of our framework. 
    For each episode, class labels are fed into the textual branch to obtain the textual embeddings.
    The support visual embeddings and query visual embeddings are extracted by the visual branch from support and query images respectively.
    To align the visual and textual embeddings, we propose a metric module to generalize the similarity measure and output the similarity score matrix.
    Moreover, for better transferability, we let the metric module can be adaptive to different few-shot tasks via bi-level optimization.
     }
    \label{fig:overall}
\end{figure}

We introduce our method of Few-shot Image classification with pre-trained Language Models (FILM) in this section.
The overall framework is illustrated in Figure~\ref{fig:overall}, which consists of three modules: a textual branch, a visual branch, and a metric module.
For each episode, the textual branch extracts textual embeddings from class labels, while the visual branch extracts visual embeddings from support and query images.
Moreover, the metric module computes the similarity score matrix between textual and visual embeddings from these two branches.
In addition, we utilize a training strategy based on MAML algorithm to train the model via bi-level optimization.

\subsection{Textual Branch}
In this section, we explain how we design the textual branch to get textual embeddings from class labels. 
The textual branch comprises a text-based pre-trained language model (PLM) and a language model head. 
During meta-training and meta-testing, the PLM is frozen while the language model head is tuned for the downstream classification tasks. 

In our study, we mainly use the masked language model as the PLM. Notice that PLMs mainly take sentences rather than single words or phrases as input during the pre-training stage.
Therefore, to bridge the gap between the pre-training and downstream tasks, for each class label $y_i$, we insert it into a hand-crafted prompt template and get $y_i^{prompt}$ as the input of the PLM. 
The token sequence of $y_i^{prompt}$ is first converted to a token embedding sequence through a token vocabulary. 
The input embedding sequence is calculated by summing the corresponding token embeddings and positional embeddings. 
Then PLM transforms the input embeddings into a sequence of hidden vectors. 
Two straightforward ways to get the textual embedding from the output hidden vector sequence are respectively: (1) taking the average vector of the output vector sequence as the textual embedding; (2) taking the hidden vector of the $\rm [CLS]$ token as the textual embedding.
To make textual embeddings more relevant to the visual descriptive information of the corresponding categories, we design a prompt template with one $\rm [MASK]$ token as 
\begin{equation*}
    y_i^{prompt} = {\rm [CLS]\enspace The\enspace appearance\enspace of\enspace }y_i {\rm\enspace is\enspace [MASK]\enspace .\enspace [SEP]}
\end{equation*}
and extract the textual embedding by sending the hidden vector of the $\rm [MASK]$ token to the language model head. 
In this way, the extraction of textual embeddings is treated as a masked language modeling task, which makes downstream classification tasks more consistent with the pre-training of the PLM.
The comparison among different designs of textual branches will be shown in Table~\ref{prompt_template} later.

\subsection{Metric Module}

Inspired by vision-language models trained by contrastive learning, we explore aligning visual and textual modalities for few-shot image classification.
However, directly aligning visual features and textual embeddings extracted by text-based PLM with cosine similarity has a poor effect in few-shot setting. 
The blue bars in Figure~\ref{fig:motativation} show that the probability of a sample image being assigned to its true label is extremely low if we directly align the visual and textual embeddings. 
In this paper, we introduce a metric module to generalize the similarity measure between visual features and textual embeddings. 
Moreover, we let the metric module adapt to different few-shot tasks for better transferability on novel classes.

Specifically, we define $f_{\theta_{I}}$ as the image encoder with learnable parameters $\theta_{I}$ to transform each sample image $x_{i}$ into a feature map $z_{i} = f_{\theta_{I}}(x_{i})$. 
Textual branch $f_{\theta_{T}}$ with learnable parameters $\theta_{T}$ is used to extract the textual embedding $t_{y_{i}} = f_{\theta_{T}}(y_{i})$ from each class label $y_{i}$.
We generalize the similarity measure between visual embeddings $z$ and textual embeddings $t$ as a learnable function $M(z, t)$ called metric module, whose parameters are denoted as $\theta_{M}$.
For example, the metric module could be a bilinear function $M(z, t)=z^\top \theta_{M}t$ (degenerating to the cosine similarity if $\theta_{M}$ is the identity matrix) or a neural network, e.g., $M(z, t)=\text{MLP}_{\theta_{M}}([z,t])$.
During meta-testing, we first fine-tune the task-specific parameters $\theta_M$ on the support set $\mathcal{S}$.
Then we use the similarity score matrix computed by the metric module as a reference to infer labels for sample images in the query set $\mathcal{Q}$.
As is shown in Figure~\ref{fig:motativation}, the correct classification probabilities of our method are significantly higher than that of direct alignment, which means that our metric module can effectively align the visual features and textual embeddings.

\subsection{Loss Function}
We formulate the learning objective as a contrastive loss (Eq~\eqref{contrastive_loss}), which pulls together images and corresponding class labels while pushing away unmatched pairs in the embedding space. 
Moreover, we aim to train a model to maximize the similarity between visual features and textual embeddings for matching (image, text) pairs while reducing the similarity for non-matching pairs.
Specifically, for a classification task $\mathcal{T}=(\mathcal{S},\mathcal{Q})$, we calculate the contrastive loss on the support set $\mathcal{S}$ and the query set $\mathcal{Q}$ respectively.
On the support set, the contrastive loss $\mathcal{L}_{\mathcal{S}}$ is computed with all the support samples, which has a formulation as:
\begin{equation}
\label{Loss_support}
\mathcal{L}_{\mathcal{S}} = 
-\frac{1}{|\mathcal{S}|}\sum_{x_i\in\mathcal{S}}\log\frac
{\exp \left( M(z_i, t_{y_i}) /\tau \right )}
{\sum_{c\in\mathcal{C}_\mathcal{T}}
\exp \left(M(z_i, t_{c})/\tau \right )},
\end{equation}
where $z_i$ is the visual embedding of the $i^{th}$ support image $x_i$, $t_{y_{i}}$ is the textual embedding of the true label $y_i$ corresponding to $x_i$, $t_c$ is the textual embedding of the class label $c$, and $M(\cdot, \cdot)$ is the similarity measure. 
On the query set, the contrastive loss $\mathcal{L}_{\mathcal{Q}}$  has almost the same formulation as $\mathcal{L}_{\mathcal{S}}$, except it is computed with all the query samples of $\mathcal{Q}$.

\subsection{Training Strategy}
In this work, we incorporate the Model-Agnostic Meta-Learning (MAML) \cite{MAML} algorithm to train the model via bi-level optimization as our training strategy. 
Our training strategy aims to learn a good model initialization (through the outer-loop optimization), which can be quickly adapted to novel tasks given a few examples (through the inner-loop optimization). 
The whole algorithm for our training strategy is outlined in Algorithm~\ref{algorithm:MAML}.

First, we randomly initialize the parameters of image encoder $\theta_I$, language model head $\theta_T$, and metric module $\theta_M$.
For each task instance $\mathcal{T}_{j}$ from the distribution $p(\mathcal{T})$, we divide $\mathcal{T}_{j}$ into a support set $\mathcal{S}_{j}$ and a query set $\mathcal{Q}_{j}$.
To let the metric module task-specific, we create copies of $\theta_M$ as the adapted parameters $\theta_M^{'}$.
In the inner loop, we adapt the model to the current task $\mathcal{T}_{j}$ by updating $\theta_M^{'}$ with a number of gradient descent steps on the support set while keeping $\theta_I$, $\theta_T$ and $\theta_M$ fixed. 
In the outer loop, $\theta_M^{'}$ are utilized to evaluate the performance of the adapted model on the query set.
Specifically, we compute loss on the query set with $\theta_I$, $\theta_T$, $\theta_{M}^{'}$ and perform gradient descent with respect to all the model parameters $\theta = \{\theta_I, \theta_T, \theta_M\}$. 
The optimization objective of the meta-training stage is to learn a good initialization across tasks.
For example, when using one gradient update in the inner loop, the optimization objective can be formulated as follows:
\begin{equation*}
\min_{\theta}  \sum_{\mathcal{T}_j \sim p(\mathcal{T})}     \mathcal{L}_{\mathcal{Q}_{j}} (\theta_I, \theta_T, \theta_M -\alpha \nabla_{\theta_{M}} \mathcal{L}_{\mathcal{S}_{j}}(\theta_I, \theta_T, \theta_M)),
\end{equation*}
where $\mathcal{L}_{\mathcal{S}_j}$ and $\mathcal{L}_{\mathcal{Q}_j}$ denote the loss functions that evaluate the performance on support and query set respectively, and $\alpha$ is the learning rate of the inner loop.

\begin{algorithm}[t]
\caption{Training strategy for our method}
\label{algorithm:MAML}
\KwIn{Task distribution $p(\mathcal{T})$, learning rate $\alpha, \beta$.}
\KwOut{Model parameters $\theta$.}

Initialize the parameters of image encoder $\theta_I$ with pre-trained model\;
Randomly initialize the parameters of language model head $\theta_T$, metric module $\theta_M$\;
\While {not done}{
 	Sample a task instance $\mathcal{T}_{j} \sim p(\mathcal{T})$\;
        Let $\mathcal{T}_{j} = (\mathcal{S}_{j}, \mathcal{Q}_{j})$ \; Initialize adapted parameters of metric module $\theta_M^{'} = \theta_M$\;
        \For{number of adaptation steps}{
            Compute loss on the support set $\mathcal{L}_{\mathcal{S}_j}(\theta_I, \theta_T, \theta_M^{'})$ using Eq \eqref{Loss_support}\;
            Update $\theta_M^{'} \leftarrow \theta_M^{'} - \alpha\nabla_{\theta_M^{'}}\mathcal{L}_{\mathcal{S}_{j}}(\theta_I, \theta_T, \theta_M^{'})$\;
            }
        Compute loss on the query set $\mathcal{L}_{\mathcal{Q}_{j}}(\theta_I, \theta_T, \theta_M^{'})$\;

    Let $\theta = \{\theta_I, \theta_T, \theta_M\}$\;
    Update $\theta \leftarrow \theta - \beta\nabla_{\theta}\mathcal{L}_{Q}(\theta_I, \theta_T, \theta_M^{'})$\;
    }

\end{algorithm}

\section{Experiments}

\subsection{Setup}
\label{experimental setup}
\textbf{Datasets.}
We experiment on three general object recognition datasets, i.e., \textit{mini}ImageNet, \textit{tiered}ImageNet and CIFAR-FS, and one fine-grained categorization image classification dataset, i.e., CUB-200-2011.
The \textit{mini}ImageNet dataset is proposed in \cite{MatchingNet} as a benchmark for few-shot image classification tasks. 
It contains a subset of 100 classes in the ImageNet \cite{ILSVRC15} dataset, where 64 classes are used for training, 16 classes for validation, and 20 classes for testing.  
The \textit{tiered}ImageNet dataset \cite{ren2018metalearning}, which is also derived from the ImageNet \cite{ILSVRC15} dataset, contains 351 classes for training, 97 classes for validation, and 160 classes for testing. 
The CIFAR-FS dataset is built upon CIFAR-100 \cite{cifar} dataset. 
Following the recent work of \cite{bertinetto2018meta}, we use the same training/validation/testing splits consisting of 64/16/20 classes respectively.
CUB-200-2011 (CUB) \cite{WahCUB_200_2011} is a dataset for fine-grained bird species classification tasks consisting of 100/50/50 classes for training/validation/testing splits respectively.
We also evaluate the domain transferability of our method by training on \textit{mini}ImageNet dataset and then testing on CUB dataset.

\textbf{Architecture.} 
For the visual branch, following previous works~\cite{oreshkin2018tadam,MetaOpt,BML}, we use ResNet-12 as our image encoder of the visual branch, which consists of four residual blocks.
Each block contains three 3$\times$3 convolutional layers and a 2$\times$2 max-pooling layer. 
Similar to~\cite{MetaOpt,BML}, we adopt Dropblock as the regularizer and set the number of filters to (64, 160, 320, 640). 
We apply a global average pooling layer after the last residual block.
The backbone network takes images with a spatial size of 84$\times$84 as input and outputs 640-dim support and query visual embeddings.
To extract comprehensive semantic information from class names, we adopt RoBERTa-base \cite{liu2019roberta} as our text-based pre-trained language model, which is trained on large-scale corpora and available for public use.
The language model is a linear layer, which transforms 768-dim hidden vectors into 640-dim textual embeddings. 
In addition, we use the bilinear form of our metric module.

\textbf{Implementation Details.} 
Following \cite{DeepBDC-CVPR2022}, we first pre-train the image encoder for 200 epochs on \emph{mini}ImageNet, CIFAR-FS and CUB dataset, and 100 epochs on \textit{tiered}ImageNet dataset.
Then we adopt the episodic training procedure under 5-way 1-shot and 5-shot settings. 
In each episode, 16 unlabeled query images per class are used for the meta-training and meta-testing phases.
We use SGD optimizer with a momentum of 0.9 and a weight decay of 5e-4. 
The outer-loop learning rate is initialized as 1e-3 on \emph{mini}ImageNet, CIFAR-FS, CUB datasets and 1e-4 on \textit{tiered}ImageNet dataset. 
The inner-loop learning rate is initialized as 0.5 on four datasets.
The number of inner-loop update steps is set to 25.
Our model is meta-trained for 80 epochs on all datasets.
The hyper-parameter $\tau$ is set as 1 for 1-shot setting, 0.2 for 5-shot setting in the inner loop, and 0.1 in the outer loop.
To ensure the stability of the evaluation results, we test 1,000 episodes and report the average performance with 95\% confidence intervals. 
We conduct experiments with an NVIDIA GeForce RTX 4090 GPU.

\begin{table*}[t]
\small
\centering
    \renewcommand\tabcolsep{7.5pt} 
    \begin{tabular}{lccccc}
    \toprule
     &  &\multicolumn{2}{c}{\emph{mini}ImageNet } &  \multicolumn{2}{c}{\emph{tiered}ImageNet }   
     \\
     Method  & Backbone & 1-shot & 5-shot & 1-shot & 5-shot  \\
    \midrule
     MAML$^\dagger$ \cite{MAML} &  ResNet-12 &  62.90$\pm$0.20 & 80.81$\pm$0.14 & 59.08$\pm$0.20 & 80.04$\pm$0.16 \\
    CC \cite{closer} & ResNet-12 & 55.43$\pm$0.81 & 77.18$\pm$0.61 & 61.49$\pm$0.91 & 82.37$\pm$0.67 
     \\
    MetaOptNet \cite{MetaOpt} & ResNet-12  & 62.64$\pm$0.61 & 78.63$\pm$0.46 & 65.99$\pm$0.72 & 81.56$\pm$0.53 
     \\
    Meta-Baseline \cite{chen2020new} & ResNet-12  & 63.17$\pm$0.23 & 79.26$\pm$0.17 & 68.62$\pm$0.27 & 83.74$\pm$0.18
     \\
    ProtoNet \cite{ProtoNet} & ResNet-12  &62.39$\pm$0.20 & 80.53$\pm$0.20 & 68.23$\pm$0.23 & 84.03$\pm$0.16 \\
    ConvNet \cite{Wertheimer2019} &ResNet-12&64.59$\pm$0.45& 82.02$\pm$0.29 &69.75$\pm$0.52&84.21$\pm$0.26
    \\
    Rethink-Distill \cite{rfs} & ResNet-12 & 64.82$\pm$0.60 & 82.14$\pm$0.43 & 71.52$\pm$0.69 & 86.03$\pm$0.49 \\
     FEAT \cite{ye2020few} & ResNet-12  & 66.78$\pm$0.20 & 82.05$\pm$0.14 & 70.80$\pm$0.23 & 84.79$\pm$0.16 
     \\
     RE-Net \cite{kang2021relational} & ResNet-12  & 67.60$\pm$0.44 & 82.58$\pm$0.30 & 71.61$\pm$0.51 & 85.28$\pm$0.35 
     \\
     CAN \cite{hou2019cross}&ResNet-12& 67.19$\pm$0.55& 80.64$\pm$0.35 &73.21$\pm$0.58&84.93$\pm$0.38
     \\
    SEMAN-G \cite{huang2022task} & ResNet-12 & 68.24$\pm$0.82 & 83.48$\pm$0.48 & 71.06$\pm$0.92 & 86.02$\pm$0.58 
     \\
     DeepEMD \cite{DeepEMD} & ResNet-12  & 65.91$\pm$0.82 & 82.41$\pm$0.56 & 71.16$\pm$0.87 & 86.03$\pm$0.58 
     \\
     TPMM \cite{wu2021task} & ResNet-12  & 67.64$\pm$0.63 & 83.44$\pm$0.43 & 72.24$\pm$0.70 & 86.55$\pm$0.63
     \\
     
	ADM \cite{ADM}   & ResNet-12 & 65.87$\pm$0.43 & 82.05$\pm$0.29&70.78$\pm$0.52 &85.70$\pm$0.43 
    \\
    \midrule
    KTN \cite{KTN} & ResNet-12  & 61.42$\pm$0.72 & 74.16$\pm$0.56 & -- & -- 
    \\
    AM3 \cite{AM3} & ResNet-12  & 65.30$\pm$0.49 & 78.10$\pm$0.36 & 69.08$\pm$0.47 & 82.58$\pm$0.31 
    \\
    TRAML \cite{TRAML} & ResNet-12  & 67.10$\pm$0.52 & 79.54$\pm$0.60 & -- & -- 
    \\
    VS-Alignment \cite{afham2022visual} & ResNet-12  & 65.89$\pm$0.80 & -- & -- & -- 
    \\
     \midrule  
    \textbf{FILM (Ours)}  & ResNet-12    & \textbf{69.52}$\pm$\textbf{0.59} & \textbf{83.68}$\pm$\textbf{0.41} & \textbf{73.28}$\pm$\textbf{0.67}& \textbf{86.72$\pm$0.45} \\    
    \bottomrule
    \end{tabular}
\caption{Comparison with previous works on \textit{mini}ImageNet and \textit{tiered}ImageNet. Results with $^\dagger$ are reported in \cite{ye2021train}. Methods in the top rows do not use semantic information, and methods in the middle rows leverage semantic information from class names \cite{AM3,KTN,TRAML} or descriptions \cite{afham2022visual}. Accuracies are reported with 95\% confidence intervals.}
\label{tab:mini-tiered}
\end{table*}

\begin{table*}[t]
\small
\centering
    \renewcommand\tabcolsep{7.5pt} 
    \begin{tabular}{lccccc}
    \toprule
     &  &\multicolumn{2}{c}{CIFAR-FS } &  \multicolumn{2}{c}{CUB-200-2011 }   
     \\
     Method  & Backbone & 1-shot & 5-shot & 1-shot & 5-shot  \\
    \midrule
    MAML~\cite{MAML} & Conv-4  &58.90$\pm$1.90  & 71.50$\pm$1.00 & 68.42$\pm$1.07 & 83.47$\pm$0.62
    \\
    CSS~\cite{AnXZZ21} & Conv-4 & 56.49$\pm$0.93  & 75.59$\pm$0.72 &66.01$\pm$0.90 & 81.84$\pm$0.59\\
    SLA~\cite{lee2020self} & Conv-4 & 48.43$\pm$0.82 &71.30$\pm$0.72 & 45.94$\pm$0.87 &68.62$\pm$0.75 \\
    CC~\cite{closer} & ResNet-12 & 60.39$\pm$0.28 & 72.85$\pm$0.65 & 67.30$\pm$0.86  & 84.75$\pm$0.60
    \\ %
    MetaOptNet~\cite{MetaOpt} & ResNet-12 & 72.60$\pm$0.70  & 84.30$\pm$0.50 & -- & --\\
    RelationNet \cite{RelationNet} & ResNet-12 &55.50$\pm$1.00  & 69.30$\pm$0.80 & 68.58$\pm$0.94& 84.05$\pm$0.56
    \\
    ProtoNet~\cite{ProtoNet} & ResNet-12 & 72.20$\pm$0.70  & 83.50$\pm$0.50 & 66.09$\pm$0.92 & 82.50$\pm$0.58
    \\ %
    R2D2 \cite{bertinetto2018meta}  & ResNet-12 & 65.30$\pm$0.02 & 78.30$\pm$0.02 & -- & -- \\
    Rethink-Distill \cite{rfs}  & ResNet-12&73.90$\pm$0.80  & 86.90$\pm$0.50 & -- & --
    \\
    RE-Net \cite{kang2021relational} & ResNet-12 & 74.51$\pm$0.46 & 86.60$\pm$0.32 & 79.49$\pm$0.44 & 91.11$\pm$0.24
    \\
    DeepEMD \cite{DeepEMD} & ResNet-12 &46.47$\pm$0.70  & 63.22$\pm$0.71 & 75.65$\pm$0.83 &  88.69$\pm$0.50
    \\
    \midrule  
    \textbf{FILM (Ours)}  &  ResNet-12 &  \textbf{77.59$\pm$0.43} &  \textbf{87.36}$\pm$\textbf{0.30} &  \textbf{79.79$\pm$0.61} &  \textbf{91.18$\pm$0.35}
     \\    
    \bottomrule
    \end{tabular}

\caption{Comparison with previous works on CIFAR-FS and CUB-200-2011. Accuracies are reported with 95\% confidence intervals.}
\label{tab:cifar-cub}
\end{table*}

\begin{table*}[t]
\centering

\resizebox{\linewidth}{!}{
\begin{minipage}{.48\linewidth}
  \centering
  \small
  \begin{tabular}{c|c}
    \toprule
    Method&\emph{mini}ImageNet $\to$ CUB
    \\ 
    \midrule
    MAML \cite{MAML}&51.34$\pm$0.72
    \\
    ProtoNet \cite{ProtoNet}&62.02$\pm$0.70
    \\
    CloserLook \cite{closer} &65.57$\pm$0.70
    \\
    Rethink-Distill \cite{rfs}&68.57$\pm$0.39
    \\ 
    Centroid \cite{afrasiyabi2020associative}&70.37$\pm$1.02
    \\
    \midrule
    \textbf{FILM (Ours)}&\textbf{71.85}$\pm$\textbf{0.54}
    \\
    \bottomrule
  \end{tabular}
\caption{Cross-domain comparison on \\ CUB dataset with 95\% confidence intervals.}
\label{Tab:mini2CUB}

\end{minipage}%
\begin{minipage}{.48\linewidth}
  \centering
  \small
    \begin{tabular}{c|ccc}
    \toprule
     Multi-Shot & 10-shot & 30-shot & 50-shot \\
    \midrule
    SimpleShot \cite{wang2019simpleshot} & 84.89  & 87.53  & 88.08  \\
    AM3 \cite{AM3}& 81.57 & -- & --
    \\
    ProtoNet \cite{ProtoNet} & 82.83  & 85.07  & 85.57 \\
    FEAT \cite{ye2020few} & 85.15   & 87.82  & 87.83 \\
    \midrule
    \textbf{FILM (Ours)} & \textbf{86.86}   & \textbf{88.92}  & \textbf{90.59} \\
    \bottomrule
  \end{tabular}
  
\caption{5-Way 10/30/50-Shot classification accuracy on {\textit{mini}ImageNet} over 1000 tasks with 95\% confidence intervals.}
\label{more shot}
\end{minipage}
}
\end{table*}

\subsection{Comparison with State-of-The-Art}
\textbf{General Object Recognition and Fine-Grained Categorization.} 
For fair comparisons, we compare with other methods using the same backbone or similar methods in both 5-way 1-shot and 5-way 5-shot settings on {\textit{mini}ImageNet}, {\textit{tiered}ImageNet}, CIFAR-FS and CUB datasets. 
As is shown in Table~\ref{tab:mini-tiered}, our method is superior to existing methods and achieves the best performance.
Compared with previous methods that leverage semantic information from class names, such as KTN \cite{KTN}, AM3 \cite{AM3}, TRAML \cite{TRAML} and Vs-Alignment \cite{VS-Alignment}, our method improves 1-shot accuracy by 2.42\% and 5-shot accuracy by 4.41\% on \textit{mini}ImageNet. 
Furthermore, our method outperforms AM3 \cite{AM3} by 3.88\% and 4.41\% at 1-shot and 5-shot settings on \textit{tiered}ImageNet respectively. 
According to Table~\ref{tab:cifar-cub}, our method outperforms MetaOptNet \cite{MetaOpt} by 4.99\% and 3.06\% at 1-shot and 5-shot settings respectively on the CIFAR-FS dataset.
In addition, on the CUB dataset, our method surpasses all the competitors, including RE-Net \cite{kang2021relational}, which previously achieved the best result.
One observation worth highlighting is that our method not only outperforms traditional methods based on meta-learning but also is superior to methods using textual information on four benchmark datasets.
These results validate the effectiveness of our proposed few-shot learning framework, which can leverage semantic information well in few-shot image classification tasks.

\textbf{Evaluation on Cross Domain and Larger Shots.} 
To evaluate the cross-domain transferability of different few-shot learning methods, we train them on the source domain \textit{mini}ImageNet dataset and test them on the target domain CUB dataset. 
This setting is challenging due to the domain gap between the training and testing datasets.
The results are reported in Table~\ref{Tab:mini2CUB}, showing that our method has competitive performance and obtains consistent improvements in the cross-domain setting.
This indicates the transferability of our method in a situation where the meta-testing tasks are entirely different from the meta-training tasks.
Furthermore, we evaluate the performance when the number of shots increases (e.g., 10-shot, 30-shot, and 50-shot) in Table~\ref{more shot}.
This shows that our method would be more effective when there are more (image, text) pairs available for novel classes.
These comparisons demonstrate that our method has a more robust transferability, which means it can work well in cross-domain and larger shots scenarios.

\begin{table*}[t]
\centering
 \renewcommand\tabcolsep{7.5pt} 
 \resizebox{0.98\textwidth}{!}{
 \begin{tabular}{c|c|c}
    \toprule
    Extraction Method & Prompt Template &\emph{mini}ImageNet 
    \\ 
    \midrule
    $\rm Avg$  & ${\rm [CLS]\enspace}y_i {\rm \enspace [SEP]}$ & 65.58$\pm$0.60
    \\
    $\rm [CLS]$ & ${\rm [CLS]\enspace}y_i {\rm \enspace [SEP]}$ & 64.93$\pm$0.61
    \\
    $\rm [CLS]$  & ${\rm [CLS]\enspace A\enspace photo\enspace of \enspace a \enspace}y_i {\rm\enspace .\enspace [SEP]}$ & 64.46$\pm$0.60
    \\
    $\rm [MASK]$ & ${\rm [CLS]\enspace A\enspace}y_i {\rm\enspace looks\enspace [MASK]\enspace .\enspace [SEP]}$ & 68.46$\pm$0.57
    \\
    \midrule
     \boldmath $\rm [MASK]$ & \boldmath${\rm [CLS]\enspace The\enspace appearance\enspace of\enspace }y_i {\rm\enspace is\enspace [MASK]\enspace .\enspace [SEP]}$&\textbf{69.52}$\pm$\textbf{0.60}
    \\
    \bottomrule
    \end{tabular}}
\caption{Comparison among different designs of the textual branch in 5-way 1-shot setting on {\textit{mini}ImageNet}.
``$\rm Avg$'' means that the textual embeddings are extracted from the average vector of all the tokens.
``$\rm [CLS]$'' means that the textual embeddings are extracted from the $\rm [CLS]$ token.
``$\rm [MASK]$'' means that the textual embeddings are extracted from the $\rm [MASK]$ token.
For all the extraction methods, we use the same PLM and language model head to extract the textual embeddings.
}
\label{prompt_template}
\end{table*}

\begin{table*}[t]
    \centering
    \renewcommand\tabcolsep{7.5pt} 
    \label{Tab:MAML}
    \resizebox{0.8\textwidth}{!}{
    \begin{tabular}{c|c|c|c|c}
    \toprule
    Metric Module &\emph{mini}ImageNet & \emph{tiered}ImageNet & CIFAR-FS & CUB-200-2011
    \\ 
    \midrule
    \XSolidBrush& 53.91$ \pm $0.68 & 51.90$ \pm $0.75 & 67.24$ \pm $0.64 & 45.97$ \pm $0.83
    \\
    \midrule
    \textbf{\Checkmark} &\textbf{69.52}$\pm$\textbf{0.60} & \textbf{72.96}$\pm$\textbf{0.48} & \textbf{77.59}$\pm$\textbf{0.43} & \textbf{79.79}$\pm$\textbf{0.61}
    \\
    \bottomrule
    \end{tabular}}
\caption{Ablation study on four widely-used benchmarks for few-shot learning. ``\XSolidBrush'' means that we remove the metric module and directly compute the cosine similarity between visual features and textual embeddings.
``\Checkmark'' means that we use our metric module to train the model.}
\label{ablation study of metric module}
\end{table*}

\subsection{Ablation Study}
In this subsection, we empirically show the effectiveness of each component. 
To investigate the effects of our designed textual branch, we try to use different extraction methods and prompt templates. 
Moreover, we conduct extensive ablation studies to verify the effectiveness in the absence of the metric module and visualize our method on \textit{mini}ImageNet and \textit{tiered}ImageNet dataset.

\textbf{Analyze of Textual Branch.}
To evaluate the effect of our textual branch, we test different extraction methods (i.e., ``Avg'', ``$\rm [CLS]$'', and ``$\rm [MASK]$'') and prompt templates in our framework with 5-way 1-shot setting on \textit{mini}ImageNet. 
As shown in Table~\ref{prompt_template}, our ``$\rm [MASK]$'' extraction method with ``${\rm [CLS]\enspace The\enspace appearance\enspace of\enspace }y_i {\rm\enspace is\enspace [MASK]\enspace .\enspace [SEP]}$'' prompt template outperforms the ``$\rm [CLS]$'' extraction method by 5.39\% and the ``$\rm Avg$'' extraction method by 3.94\%. 
Our proposed hand-crafted prompt template treats the extraction of textual embeddings as a masked language modeling task, which makes the textual embeddings more relevant to the visual description of object categories.
The results demonstrate that the carefully designed textual branch is effective for aligning visual and textual embeddings for downstream few-shot classification tasks.

\textbf{Analyze of Metric Module.} 
As is shown in Table~\ref{ablation study of metric module}, we design a new model without using the support set to update the parameters in the inner-loop optimization and directly compute the similarity score matrix between the query visual embeddings and textual embeddings with cosine similarity in the outer loop. 
The results show a significant decrease in performance on four widely-used few-shot image classification datasets, demonstrating the importance of the task-specific metric module.
By leveraging the metric module to generalize the cosine similarity, our model can adaptively measure the similarity between visual features and textual embeddings for different few-shot tasks.

\begin{figure}
    \centering
    \includegraphics[width=0.7\textwidth]{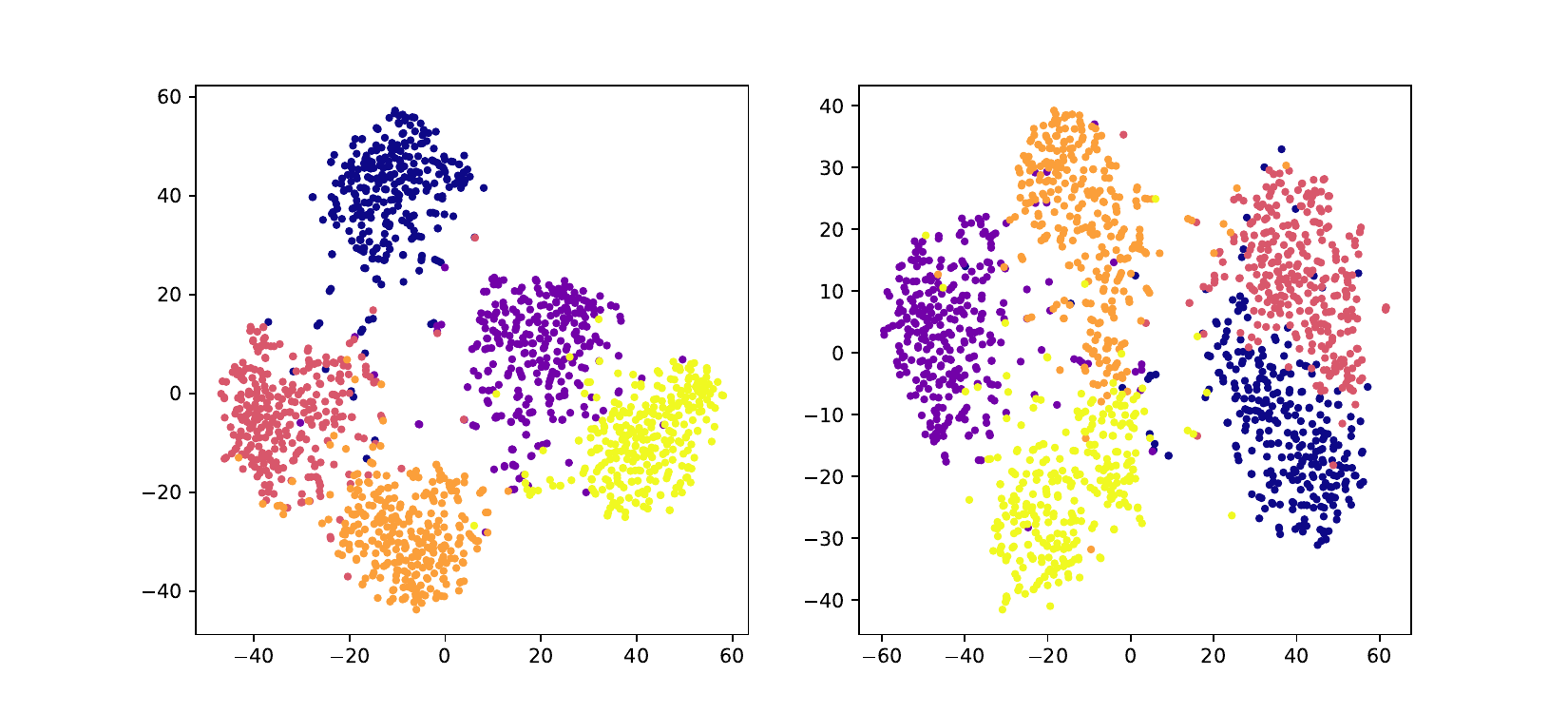}
    \caption{$t$-SNE visualization of the distribution of our method with 5-way setting on \textit{mini}ImageNet dataset (left) and \textit{tiered}ImageNet dataset (right). Dots in different colors stand for visual embeddings of different categories.}
    \label{fig:t-SNE}
\end{figure}

\textbf{Visualization.} 
To qualitatively evaluate our method, we apply $t$-SNE \cite{LaurensVisualizing2008} to visualize the results, which represent the visual features of five categories.
We randomly sample 300 examples for each class in 5-way 5-shot setting on \textit{mini}ImageNet and \textit{tiered}ImageNet dataset.
As shown in Figure~\ref{fig:t-SNE}, the $t$-SNE visualization results indicate that our method can learn more compact and separate clusters, which means that the learned representations are more discriminative.

\section{Conclusion}
In this paper, we propose a novel few-shot learning framework with text-based pre-trained language model to boost few-shot learning.
Furthermore, we introduce a task-specific metric module to enable the alignment between visual features and textual embeddings.
Extensive experiments on \textit{mini}ImageNet, \textit{tiered}ImageNet and CIFAR-FS demonstrate the effectiveness of our method.

{\bibliography{reference}

\begin{thebibliography}{67}
\providecommand{\natexlab}[1]{#1}
\providecommand{\url}[1]{\texttt{#1}}
\expandafter\ifx\csname urlstyle\endcsname\relax
  \providecommand{\doi}[1]{doi: #1}\else
  \providecommand{\doi}{doi: \begingroup \urlstyle{rm}\Url}\fi

\bibitem[Krizhevsky et~al.(2017)Krizhevsky, Sutskever, and Hinton]{AlexNet}
Alex Krizhevsky, Ilya Sutskever, and Geoffrey~E Hinton.
\newblock Imagenet classification with deep convolutional neural networks.
\newblock \emph{Communications of the ACM}, 60\penalty0 (6):\penalty0 84--90,
  2017.

\bibitem[Simonyan and Zisserman(2014)]{VGGNet}
Karen Simonyan and Andrew Zisserman.
\newblock Very deep convolutional networks for large-scale image recognition.
\newblock \emph{arXiv preprint arXiv:1409.1556}, 2014.

\bibitem[Szegedy et~al.(2015)Szegedy, Liu, Jia, Sermanet, Reed, Anguelov,
  Erhan, Vanhoucke, and Rabinovich]{GoogLeNet}
Christian Szegedy, Wei Liu, Yangqing Jia, Pierre Sermanet, Scott Reed, Dragomir
  Anguelov, Dumitru Erhan, Vincent Vanhoucke, and Andrew Rabinovich.
\newblock Going deeper with convolutions.
\newblock In \emph{Proceedings of the IEEE conference on computer vision and
  pattern recognition}, pages 1--9, 2015.

\bibitem[He et~al.(2016)He, Zhang, Ren, and Sun]{ResNet}
Kaiming He, Xiangyu Zhang, Shaoqing Ren, and Jian Sun.
\newblock Deep residual learning for image recognition.
\newblock In \emph{Proceedings of the IEEE conference on computer vision and
  pattern recognition}, pages 770--778, 2016.

\bibitem[Bart and Ullman(2005)]{FSL1}
Evgeniy Bart and Shimon Ullman.
\newblock Cross-generalization: Learning novel classes from a single example by
  feature replacement.
\newblock In \emph{2005 IEEE Computer Society Conference on Computer Vision and
  Pattern Recognition (CVPR'05)}, volume~1, pages 672--679. IEEE, 2005.

\bibitem[Fink(2004)]{FSL2}
Michael Fink.
\newblock Object classification from a single example utilizing class relevance
  metrics.
\newblock \emph{Advances in neural information processing systems}, 17, 2004.

\bibitem[Fei-Fei et~al.(2006)Fei-Fei, Fergus, and Perona]{FSL3}
Li~Fei-Fei, Robert Fergus, and Pietro Perona.
\newblock One-shot learning of object categories.
\newblock \emph{IEEE transactions on pattern analysis and machine
  intelligence}, 28\penalty0 (4):\penalty0 594--611, 2006.

\bibitem[Lake et~al.(2011)Lake, Salakhutdinov, Gross, and Tenenbaum]{FSL4}
Brenden Lake, Ruslan Salakhutdinov, Jason Gross, and Joshua Tenenbaum.
\newblock One shot learning of simple visual concepts.
\newblock In \emph{Proceedings of the annual meeting of the cognitive science
  society}, volume~33, 2011.

\bibitem[Vinyals et~al.(2016)Vinyals, Blundell, Lillicrap, Wierstra,
  et~al.]{MatchingNet}
Oriol Vinyals, Charles Blundell, Timothy Lillicrap, Daan Wierstra, et~al.
\newblock Matching networks for one shot learning.
\newblock \emph{Advances in neural information processing systems}, 29, 2016.

\bibitem[Snell et~al.(2017)Snell, Swersky, and Zemel]{ProtoNet}
Jake Snell, Kevin Swersky, and Richard Zemel.
\newblock Prototypical networks for few-shot learning.
\newblock \emph{Advances in neural information processing systems}, 30, 2017.

\bibitem[Finn et~al.(2017)Finn, Abbeel, and Levine]{MAML}
Chelsea Finn, Pieter Abbeel, and Sergey Levine.
\newblock Model-agnostic meta-learning for fast adaptation of deep networks.
\newblock In \emph{ICML}, 2017.

\bibitem[Sung et~al.(2018)Sung, Yang, Zhang, Xiang, Torr, and
  Hospedales]{RelationNet}
Flood Sung, Yongxin Yang, Li~Zhang, Tao Xiang, Philip~HS Torr, and Timothy~M
  Hospedales.
\newblock Learning to compare: Relation network for few-shot learning.
\newblock In \emph{Proceedings of the IEEE conference on computer vision and
  pattern recognition}, pages 1199--1208, 2018.

\bibitem[et~al(2020)]{DeepEMD}
Zhang et~al.
\newblock Deepemd: Few-shot image classification with differentiable earth
  mover's distance and structured classifiers.
\newblock In \emph{ICCV}, 2020.

\bibitem[Xing et~al.(2019)Xing, Rostamzadeh, Oreshkin, and O~Pinheiro]{AM3}
Chen Xing, Negar Rostamzadeh, Boris Oreshkin, and Pedro~O O~Pinheiro.
\newblock Adaptive cross-modal few-shot learning.
\newblock \emph{Advances in Neural Information Processing Systems}, 32, 2019.

\bibitem[Peng et~al.(2019)Peng, Li, Zhang, Li, Qi, and Tang]{KTN}
Zhimao Peng, Zechao Li, Junge Zhang, Yan Li, Guo-Jun Qi, and Jinhui Tang.
\newblock Few-shot image recognition with knowledge transfer.
\newblock In \emph{Proceedings of the IEEE/CVF International Conference on
  Computer Vision}, pages 441--449, 2019.

\bibitem[Li et~al.(2020{\natexlab{a}})Li, Huang, Lan, Feng, Li, and
  Wang]{TRAML}
Aoxue Li, Weiran Huang, Xu~Lan, Jiashi Feng, Zhenguo Li, and Liwei Wang.
\newblock Boosting few-shot learning with adaptive margin loss.
\newblock In \emph{Proceedings of the IEEE/CVF conference on computer vision
  and pattern recognition}, pages 12576--12584, 2020{\natexlab{a}}.

\bibitem[Pennington et~al.(2014)Pennington, Socher, and Manning]{GloVe}
Jeffrey Pennington, Richard Socher, and Christopher~D Manning.
\newblock Glove: Global vectors for word representation.
\newblock In \emph{Proceedings of the 2014 conference on empirical methods in
  natural language processing (EMNLP)}, pages 1532--1543, 2014.

\bibitem[Devlin et~al.(2018)Devlin, Chang, Lee, and Toutanova]{BERT}
Jacob Devlin, Ming-Wei Chang, Kenton Lee, and Kristina Toutanova.
\newblock Bert: Pre-training of deep bidirectional transformers for language
  understanding.
\newblock \emph{arXiv preprint arXiv:1810.04805}, 2018.

\bibitem[Radford et~al.(2018)Radford, Narasimhan, Salimans, Sutskever,
  et~al.]{GPT}
Alec Radford, Karthik Narasimhan, Tim Salimans, Ilya Sutskever, et~al.
\newblock Improving language understanding by generative pre-training.
\newblock 2018.

\bibitem[Afham and Rodrigo(2022{\natexlab{a}})]{VS-Alignment}
Mohamed Afham and Ranga Rodrigo.
\newblock Visual-semantic contrastive alignment for few-shot image
  classification.
\newblock \emph{arXiv preprint arXiv:2210.11000}, 2022{\natexlab{a}}.

\bibitem[Chen et~al.(2023)Chen, Si, Zhang, Wang, Wang, and Tan]{SP}
Wentao Chen, Chenyang Si, Zhang Zhang, Liang Wang, Zilei Wang, and Tieniu Tan.
\newblock Semantic prompt for few-shot image recognition.
\newblock \emph{arXiv preprint arXiv:2303.14123}, 2023.

\bibitem[Radford et~al.(2021)Radford, Kim, Hallacy, Ramesh, Goh, Agarwal,
  Sastry, Askell, Mishkin, Clark, et~al.]{CLIP}
Alec Radford, Jong~Wook Kim, Chris Hallacy, Aditya Ramesh, Gabriel Goh,
  Sandhini Agarwal, Girish Sastry, Amanda Askell, Pamela Mishkin, Jack Clark,
  et~al.
\newblock Learning transferable visual models from natural language
  supervision.
\newblock In \emph{International Conference on Machine Learning}, pages
  8748--8763. PMLR, 2021.

\bibitem[Jia et~al.(2021)Jia, Yang, Xia, Chen, Parekh, Pham, Le, Sung, Li, and
  Duerig]{ALIGN}
Chao Jia, Yinfei Yang, Ye~Xia, Yi-Ting Chen, Zarana Parekh, Hieu Pham, Quoc Le,
  Yun-Hsuan Sung, Zhen Li, and Tom Duerig.
\newblock Scaling up visual and vision-language representation learning with
  noisy text supervision.
\newblock In \emph{International Conference on Machine Learning}, pages
  4904--4916. PMLR, 2021.

\bibitem[Yoon et~al.(2019)Yoon, Seo, and Moon]{Tapnet}
Sung~Whan Yoon, Jun Seo, and Jaekyun Moon.
\newblock Tapnet: Neural network augmented with task-adaptive projection for
  few-shot learning.
\newblock In \emph{International conference on machine learning}, pages
  7115--7123. PMLR, 2019.

\bibitem[Li et~al.(2019)Li, Eigen, Dodge, Zeiler, and Wang]{CTM}
Hongyang Li, David Eigen, Samuel Dodge, Matthew Zeiler, and Xiaogang Wang.
\newblock {Finding Task-Relevant Features for Few-Shot Learning by Category
  Traversal}.
\newblock In \emph{CVPR}, 2019.

\bibitem[Qiao et~al.(2019)Qiao, Shi, Li, Wang, Huang, and Tian]{TEAM}
Limeng Qiao, Yemin Shi, Jia Li, Yaowei Wang, Tiejun Huang, and Yonghong Tian.
\newblock Transductive episodic-wise adaptive metric for few-shot learning.
\newblock In \emph{Proceedings of the IEEE/CVF international conference on
  computer vision}, pages 3603--3612, 2019.

\bibitem[Ye et~al.(2020)Ye, Hu, Zhan, and Sha]{ye2020few}
Han-Jia Ye, Hexiang Hu, De-Chuan Zhan, and Fei Sha.
\newblock Few-shot learning via embedding adaptation with set-to-set functions.
\newblock In \emph{Proceedings of the IEEE/CVF conference on computer vision
  and pattern recognition}, pages 8808--8817, 2020.

\bibitem[Simon et~al.(2020)Simon, Koniusz, Nock, and Harandi]{DSN}
Christian Simon, Piotr Koniusz, Richard Nock, and Mehrtash Harandi.
\newblock Adaptive subspaces for few-shot learning.
\newblock In \emph{Proceedings of the IEEE/CVF conference on computer vision
  and pattern recognition}, pages 4136--4145, 2020.

\bibitem[Rusu et~al.(2018)Rusu, Rao, Sygnowski, Vinyals, Pascanu, Osindero, and
  Hadsell]{LEO}
Andrei~A Rusu, Dushyant Rao, Jakub Sygnowski, Oriol Vinyals, Razvan Pascanu,
  Simon Osindero, and Raia Hadsell.
\newblock Meta-learning with latent embedding optimization.
\newblock \emph{arXiv preprint arXiv:1807.05960}, 2018.

\bibitem[Sun et~al.(2019)Sun, Liu, Chua, and Schiele]{MTL}
Qianru Sun, Yaoyao Liu, Tat-Seng Chua, and Bernt Schiele.
\newblock Meta-transfer learning for few-shot learning.
\newblock In \emph{Proceedings of the IEEE/CVF Conference on Computer Vision
  and Pattern Recognition}, pages 403--412, 2019.

\bibitem[Lee et~al.(2019)Lee, Maji, Ravichandran, and Soatto]{MetaOpt}
Kwonjoon Lee, Subhransu Maji, Avinash Ravichandran, and Stefano Soatto.
\newblock Meta-learning with differentiable convex optimization.
\newblock In \emph{Proceedings of the IEEE/CVF Conference on Computer Vision
  and Pattern Recognition}, pages 10657--10665, 2019.

\bibitem[Kipf and Welling(2016)]{GCN}
Thomas~N Kipf and Max Welling.
\newblock Semi-supervised classification with graph convolutional networks.
\newblock \emph{arXiv preprint arXiv:1609.02907}, 2016.

\bibitem[Dosovitskiy et~al.(2020)Dosovitskiy, Beyer, Kolesnikov, Weissenborn,
  Zhai, Unterthiner, Dehghani, Minderer, Heigold, Gelly, et~al.]{ViT}
Alexey Dosovitskiy, Lucas Beyer, Alexander Kolesnikov, Dirk Weissenborn,
  Xiaohua Zhai, Thomas Unterthiner, Mostafa Dehghani, Matthias Minderer, Georg
  Heigold, Sylvain Gelly, et~al.
\newblock An image is worth 16x16 words: Transformers for image recognition at
  scale.
\newblock \emph{arXiv preprint arXiv:2010.11929}, 2020.

\bibitem[He et~al.(2020)He, Fan, Wu, Xie, and Girshick]{MoCo}
Kaiming He, Haoqi Fan, Yuxin Wu, Saining Xie, and Ross Girshick.
\newblock Momentum contrast for unsupervised visual representation learning.
\newblock In \emph{Proceedings of the IEEE/CVF conference on computer vision
  and pattern recognition}, pages 9729--9738, 2020.

\bibitem[Chen et~al.(2020{\natexlab{a}})Chen, Fan, Girshick, and He]{MoCov2}
Xinlei Chen, Haoqi Fan, Ross Girshick, and Kaiming He.
\newblock Improved baselines with momentum contrastive learning.
\newblock \emph{arXiv preprint arXiv:2003.04297}, 2020{\natexlab{a}}.

\bibitem[Chen et~al.(2020{\natexlab{b}})Chen, Kornblith, Norouzi, and
  Hinton]{SimCLR}
Ting Chen, Simon Kornblith, Mohammad Norouzi, and Geoffrey Hinton.
\newblock A simple framework for contrastive learning of visual
  representations.
\newblock In \emph{International conference on machine learning}, pages
  1597--1607. PMLR, 2020{\natexlab{b}}.

\bibitem[Liu and Sun(2015)]{WordAlignment}
Yang Liu and Maosong Sun.
\newblock Contrastive unsupervised word alignment with non-local features.
\newblock In \emph{Proceedings of the AAAI Conference on Artificial
  Intelligence}, volume~29, 2015.

\bibitem[Huang et~al.(2018)Huang, Li, Ping, and Huang]{LanguageModeling}
Jiaji Huang, Yi~Li, Wei Ping, and Liang Huang.
\newblock Large margin neural language model.
\newblock \emph{arXiv preprint arXiv:1808.08987}, 2018.

\bibitem[Lee et~al.(2020{\natexlab{a}})Lee, Lee, and Hwang]{MachineTranslation}
Seanie Lee, Dong~Bok Lee, and Sung~Ju Hwang.
\newblock Contrastive learning with adversarial perturbations for conditional
  text generation.
\newblock \emph{arXiv preprint arXiv:2012.07280}, 2020{\natexlab{a}}.

\bibitem[Zhang et~al.(2022)Zhang, Jiang, Miura, Manning, and Langlotz]{ConVIRT}
Yuhao Zhang, Hang Jiang, Yasuhide Miura, Christopher~D Manning, and Curtis~P
  Langlotz.
\newblock Contrastive learning of medical visual representations from paired
  images and text.
\newblock In \emph{Machine Learning for Healthcare Conference}, pages 2--25.
  PMLR, 2022.

\bibitem[Alayrac et~al.(2022)Alayrac, Donahue, Luc, Miech, Barr, Hasson, Lenc,
  Mensch, Millican, Reynolds, et~al.]{Flamingo}
Jean-Baptiste Alayrac, Jeff Donahue, Pauline Luc, Antoine Miech, Iain Barr,
  Yana Hasson, Karel Lenc, Arthur Mensch, Katherine Millican, Malcolm Reynolds,
  et~al.
\newblock Flamingo: a visual language model for few-shot learning.
\newblock \emph{Advances in Neural Information Processing Systems},
  35:\penalty0 23716--23736, 2022.

\bibitem[Oord et~al.(2018)Oord, Li, and Vinyals]{oord2018representation}
Aaron van~den Oord, Yazhe Li, and Oriol Vinyals.
\newblock Representation learning with contrastive predictive coding.
\newblock \emph{arXiv preprint arXiv:1807.03748}, 2018.

\bibitem[Russakovsky et~al.(2015)Russakovsky, Deng, Su, Krause, Satheesh, Ma,
  Huang, Karpathy, Khosla, Bernstein, Berg, and Fei-Fei]{ILSVRC15}
Olga Russakovsky, Jia Deng, Hao Su, Jonathan Krause, Sanjeev Satheesh, Sean Ma,
  Zhiheng Huang, Andrej Karpathy, Aditya Khosla, Michael Bernstein,
  Alexander~C. Berg, and Li~Fei-Fei.
\newblock {ImageNet Large Scale Visual Recognition Challenge}.
\newblock \emph{IJCV}, 2015.

\bibitem[Ren et~al.(2018)Ren, Triantafillou, Ravi, Snell, Swersky, Tenenbaum,
  Larochelle, and Zemel]{ren2018metalearning}
Mengye Ren, Eleni Triantafillou, Sachin Ravi, Jake Snell, Kevin Swersky,
  Joshua~B. Tenenbaum, Hugo Larochelle, and Richard~S. Zemel.
\newblock Meta-learning for semi-supervised few-shot classification.
\newblock In \emph{ICLR}, 2018.

\bibitem[Krizhevsky et~al.(2009)Krizhevsky, Hinton, et~al.]{cifar}
Alex Krizhevsky, Geoffrey Hinton, et~al.
\newblock Learning multiple layers of features from tiny images.
\newblock 2009.

\bibitem[Bertinetto et~al.(2018)Bertinetto, Henriques, Torr, and
  Vedaldi]{bertinetto2018meta}
Luca Bertinetto, Joao~F Henriques, Philip~HS Torr, and Andrea Vedaldi.
\newblock Meta-learning with differentiable closed-form solvers.
\newblock \emph{arXiv preprint arXiv:1805.08136}, 2018.

\bibitem[Wah et~al.(2011)Wah, Branson, Welinder, Perona, and
  Belongie]{WahCUB_200_2011}
C.~Wah, S.~Branson, P.~Welinder, P.~Perona, and S.~Belongie.
\newblock {The Caltech-UCSD Birds-200-2011 Dataset}.
\newblock Technical Report CNS-TR-2011-001, California Institute of Technology,
  2011.

\bibitem[Oreshkin et~al.(2018)Oreshkin, Rodr{\'\i}guez~L{\'o}pez, and
  Lacoste]{oreshkin2018tadam}
Boris Oreshkin, Pau Rodr{\'\i}guez~L{\'o}pez, and Alexandre Lacoste.
\newblock Tadam: Task dependent adaptive metric for improved few-shot learning.
\newblock \emph{Advances in neural information processing systems}, 31, 2018.

\bibitem[Zhou et~al.(2021)Zhou, Qiu, Xie, Wu, and Zhang]{BML}
Ziqi Zhou, Xi~Qiu, Jiangtao Xie, Jianan Wu, and Chi Zhang.
\newblock Binocular mutual learning for improving few-shot classification.
\newblock In \emph{Proceedings of the IEEE/CVF International Conference on
  Computer Vision}, pages 8402--8411, 2021.

\bibitem[Liu et~al.(2019)Liu, Ott, Goyal, Du, Joshi, Chen, Levy, Lewis,
  Zettlemoyer, and Stoyanov]{liu2019roberta}
Yinhan Liu, Myle Ott, Naman Goyal, Jingfei Du, Mandar Joshi, Danqi Chen, Omer
  Levy, Mike Lewis, Luke Zettlemoyer, and Veselin Stoyanov.
\newblock Roberta: A robustly optimized bert pretraining approach.
\newblock \emph{arXiv preprint arXiv:1907.11692}, 2019.

\bibitem[Xie et~al.(2022)Xie, Long, Lv, Wang, and Li]{DeepBDC-CVPR2022}
Jiangtao Xie, Fei Long, Jiaming Lv, Qilong Wang, and Peihua Li.
\newblock Joint distribution matters: Deep brownian distance covariance for
  few-shot classification.
\newblock In \emph{CVPR}, 2022.

\bibitem[Chen et~al.(2019)Chen, Liu, Kira, Wang, and Huang]{closer}
Wei-Yu Chen, Yen-Cheng Liu, Zsolt Kira, Yu-Chiang Wang, and Jia-Bin Huang.
\newblock A closer look at few-shot classification.
\newblock In \emph{International Conference on Learning Representations}, 2019.

\bibitem[Chen et~al.(2021)Chen, Liu, Xu, Darrell, and Wang]{chen2020new}
Yinbo Chen, Zhuang Liu, Huijuan Xu, Trevor Darrell, and Xiaolong Wang.
\newblock Meta-baseline: exploring simple meta-learning for few-shot learning.
\newblock In \emph{Proceedings of the IEEE/CVF International Conference on
  Computer Vision}, pages 9062--9071, 2021.

\bibitem[Wertheimer and Hariharan(2019)]{Wertheimer2019}
Davis Wertheimer and Bharath Hariharan.
\newblock Few-shot learning with localization in realistic settings.
\newblock In \emph{CVPR}, 2019.

\bibitem[Tian et~al.(2020)Tian, Wang, Krishnan, Tenenbaum, and Isola]{rfs}
Yonglong Tian, Yue Wang, Dilip Krishnan, Joshua~B Tenenbaum, and Phillip Isola.
\newblock Rethinking few-shot image classification: a good embedding is all you
  need?
\newblock In \emph{Computer Vision--ECCV 2020: 16th European Conference}.
  Springer, 2020.

\bibitem[Kang et~al.(2021)Kang, Kwon, Min, and Cho]{kang2021relational}
Dahyun Kang, Heeseung Kwon, Juhong Min, and Minsu Cho.
\newblock Relational embedding for few-shot classification.
\newblock In \emph{Proceedings of the IEEE/CVF International Conference on
  Computer Vision}, pages 8822--8833, 2021.

\bibitem[Hou et~al.(2019)Hou, Chang, Ma, Shan, and Chen]{hou2019cross}
Ruibing Hou, Hong Chang, Bingpeng Ma, Shiguang Shan, and Xilin Chen.
\newblock Cross attention network for few-shot classification.
\newblock \emph{Advances in Neural Information Processing Systems}, 32, 2019.

\bibitem[Huang et~al.(2022)Huang, Ma, Han, and Chang]{huang2022task}
Shiyuan Huang, Jiawei Ma, Guangxing Han, and Shih-Fu Chang.
\newblock Task-adaptive negative envision for few-shot open-set recognition.
\newblock In \emph{Proceedings of the IEEE/CVF Conference on Computer Vision
  and Pattern Recognition}, pages 7171--7180, 2022.

\bibitem[Wu et~al.(2021)Wu, Zhang, Zhang, and Wu]{wu2021task}
Jiamin Wu, Tianzhu Zhang, Yongdong Zhang, and Feng Wu.
\newblock Task-aware part mining network for few-shot learning.
\newblock In \emph{Proceedings of the IEEE/CVF International Conference on
  Computer Vision}, pages 8433--8442, 2021.

\bibitem[Li et~al.(2020{\natexlab{b}})Li, Wang, Huo, Shi, Gao, and Luo]{ADM}
Wenbin Li, Lei Wang, Jing Huo, Yinghuan Shi, Yang Gao, and Jiebo Luo.
\newblock {Asymmetric distribution measure for few-shot learning}.
\newblock \emph{IJCAI}, 2020{\natexlab{b}}.

\bibitem[Afham and Rodrigo(2022{\natexlab{b}})]{afham2022visual}
Mohamed Afham and Ranga Rodrigo.
\newblock Visual-semantic contrastive alignment for few-shot image
  classification.
\newblock \emph{arXiv preprint arXiv:2210.11000}, 2022{\natexlab{b}}.

\bibitem[Ye and Chao(2021)]{ye2021train}
Han-Jia Ye and Wei-Lun Chao.
\newblock How to train your maml to excel in few-shot classification.
\newblock \emph{arXiv preprint arXiv:2106.16245}, 2021.

\bibitem[An et~al.(2021)An, Xue, Zhao, and Zhang]{AnXZZ21}
Yuexuan An, Hui Xue, Xingyu Zhao, and Lu~Zhang.
\newblock Conditional self-supervised learning for few-shot classification.
\newblock In \emph{International Joint Conference on Artificial Intelligence,
  {IJCAI}}, 2021.

\bibitem[Lee et~al.(2020{\natexlab{b}})Lee, Hwang, and Shin]{lee2020self}
Hankook Lee, Sung~Ju Hwang, and Jinwoo Shin.
\newblock Self-supervised label augmentation via input transformations.
\newblock In \emph{International Conference on Machine Learning,{ICML}},
  2020{\natexlab{b}}.

\bibitem[Afrasiyabi et~al.(2020)Afrasiyabi, Lalonde, and
  Gagn{\'e}]{afrasiyabi2020associative}
Arman Afrasiyabi, Jean-Franccois Lalonde, and Christian Gagn{\'e}.
\newblock Associative alignment for few-shot image classification.
\newblock In \emph{ECCV}, 2020.

\bibitem[Wang et~al.(2019)Wang, Chao, Weinberger, and van~der
  Maaten]{wang2019simpleshot}
Yan Wang, Wei-Lun Chao, Kilian~Q Weinberger, and Laurens van~der Maaten.
\newblock Simpleshot: Revisiting nearest-neighbor classification for few-shot
  learning.
\newblock \emph{CoRR}, abs/1911.04623, 2019.

\bibitem[Laurens et~al.(2008)Laurens, Maaten, Hinton, and
  Geoffrey]{LaurensVisualizing2008}
Laurens, Van~Der Maaten, Hinton, and Geoffrey.
\newblock Visualizing data using t-sne.
\newblock \emph{Journal of Machine Learning Research}, 2008.

\end{thebibliography}
\bibliographystyle{unsrtnat}}

\clearpage
\appendix
\begin{center}
    {\LARGE Supplementary Materials} 
\end{center}

\section{Additional Experiments}

\textbf{Influence of Inner-Loop Temperature.}
To study the influence of inner-loop temperature hyper-parameter, we conduct experiments on four widely-used few-shot datasets with different inner-loop temperature values in our method.
The rest settings are consistent with Section~\ref{experimental setup}.
Table~\ref{tab:temperature} shows the results in 5-way 5-shot setting.
We find that 0.2 is an appropriate inner-loop temperature value for this setting on all these four datasets.

\begin{table}[h]
    \centering
    \renewcommand\tabcolsep{7.5pt} 
    \label{Tab:Tempeature}
    \resizebox{0.95\textwidth}{!}{
    \begin{tabular}{c|c|c|c|c}
    \toprule
    Inner-Loop Temperature &\emph{mini}ImageNet & \emph{tiered}ImageNet & CIFAR-FS & CUB-200-2011
    \\ 
    \midrule
    1   & 77.56$ \pm $0.50 & 78.14$ \pm $0.63 &  80.19$ \pm $0.57 & 85.38$ \pm $0.50
    \\
    0.7  & 80.51$ \pm $0.46 & 81.70$ \pm $0.57 & 84.35$ \pm $0.52 &  88.46$ \pm $0.44
    \\
    0.5  & 82.42$ \pm $0.43 & 83.74$ \pm $0.54 & 86.13$ \pm $0.49 & 89.56$ \pm $0.41
    \\
    0.3  & 83.49$ \pm $0.41 & 86.33$ \pm $0.49 & 87.31$ \pm $0.45 & 90.85$ \pm $0.36
    \\
    \textbf{0.2}  & \textbf{83.68$ \pm $0.41} & \textbf{86.72$ \pm $0.45} & \textbf{87.36$ \pm $0.31} & \textbf{91.18$ \pm $0.35}
    \\
    0.1  & 82.11$ \pm $0.41 & 86.24$ \pm $0.48 & 87.12$ \pm $0.44 & 90.42$ \pm $0.38
    \\
    \bottomrule
    \end{tabular}}
    \vspace{10pt}
\caption{Ablation studies on the inner-loop temperature.}
\label{tab:temperature}
\end{table}

\textbf{Effect of the Number of Inner-Loop Update Steps.}
To find a suitable number of inner-loop update steps, we keep the experimental setup in Section~\ref{experimental setup} and update the model 10, 15, 20, 25 and 30 steps in the inner loop respectively.
Table~\ref{tab:step} shows the results in 5-way 5-shot setting on \emph{mini}ImageNet and \emph{tiered}ImageNet. 
Following the results, we set the number of inner-loop update steps to 25 in our experiments.

\begin{table}[h]
    \centering
    \renewcommand\tabcolsep{7.5pt} 
    \label{Tab:step}
    \resizebox{0.9\textwidth}{!}{
    \begin{tabular}{c|c|c|c|c|c}
    \toprule
    Number of Steps  & 10 & 15 & 20 & \textbf{25} & 30 
    \\ 
    \midrule
     \emph{mini}ImageNet  & 82.59$ \pm $0.42 & 83.20$ \pm $0.40 & 83.28$ \pm $0.40 & \textbf{83.68$ \pm $0.67} & 83.10$ \pm $0.41
     \\
     \emph{tiered}ImageNet  & 85.20$ \pm $0.51 & 86.14$ \pm $0.49 & 86.02$ \pm $0.33 & \textbf{86.73$ \pm $0.45} & 86.61$ \pm $0.48
     \\
    \bottomrule
    \end{tabular}}
    \vspace{10pt}
\caption{Ablation studies on the number of inner-loop update steps.}
\label{tab:step}
\end{table}

\begin{figure}[h]
    \centering
    \includegraphics[width=1\textwidth]{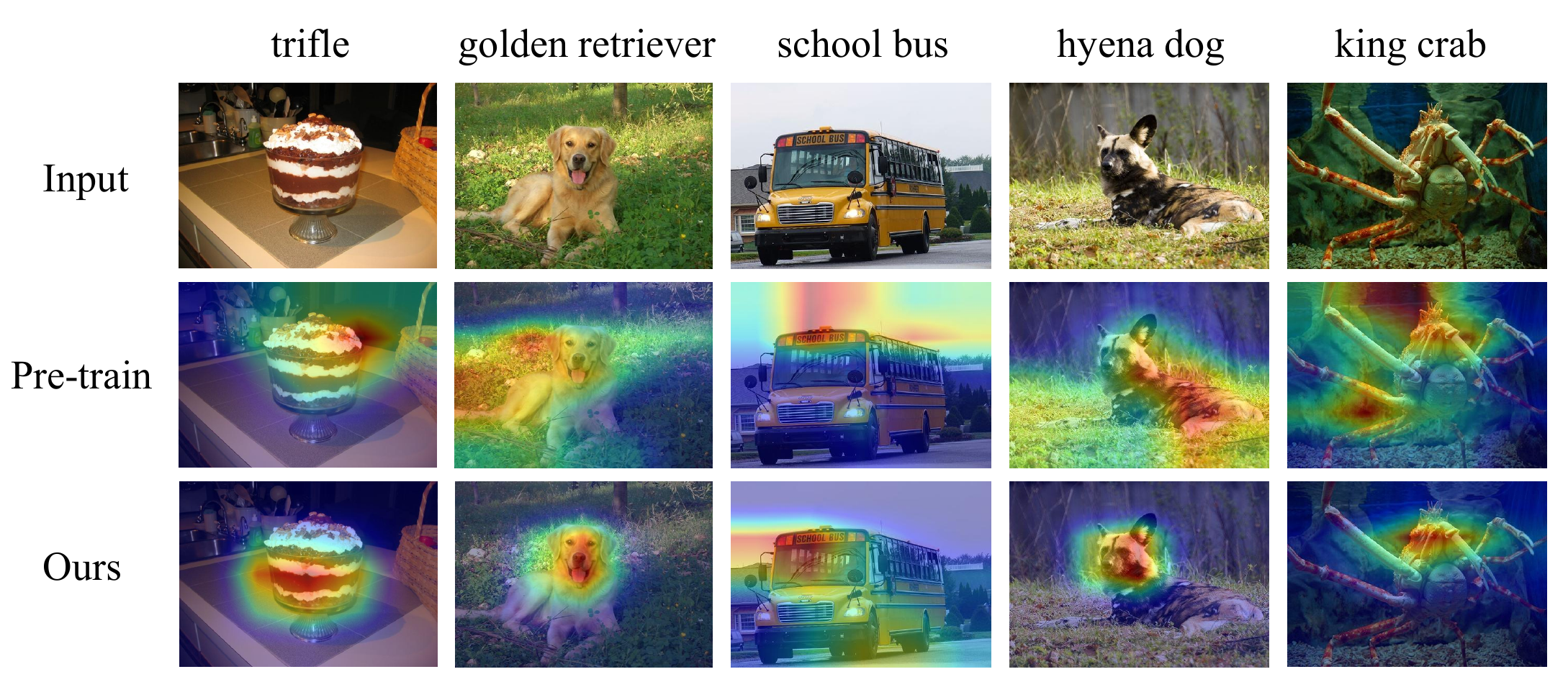}
    \caption{Grad-CAM visualization of \emph{mini}ImageNet dataset.}
    \label{fig:gradcam}
\end{figure}

\textbf{Visualization of Grad-CAM.} 
In Figure~\ref{fig:gradcam}, we visualize the gradient-weighted class activation mapping from the pre-trained model and our method under a ResNet-12 feature extractor.
It is observed that our method makes the model pay more attention to the discriminative part of the target object than the pre-trained model.
For example, we find that for dog samples, the pre-trained model pays more attention to the body and background parts while our model focuses on the head part.

\end{document}